\documentclass[letterpaper, 10 pt, conference]{ieeeconf}  

\IEEEoverridecommandlockouts                              

\usepackage[utf8]{inputenc} 
\usepackage[T1]{fontenc}    
\usepackage{hyperref}       
\usepackage{url}            
\usepackage{booktabs}       
\usepackage{amsfonts}       
\usepackage{nicefrac}       
\usepackage{microtype}      
\usepackage{xcolor}         
\usepackage{wrapfig}
\usepackage{subcaption}
\usepackage{multirow}
\usepackage{microtype}
\usepackage{graphicx}
\usepackage{color}
\usepackage{comment}
\usepackage{amsmath}
\usepackage{amssymb}
\let\labelindent\relax
\usepackage{enumitem}
\usepackage{cleveref}
\Crefname{equation}{Eq.}{Eqs.}
\Crefname{figure}{Fig.}{Figs.}
\usepackage{color}
\usepackage{balance}

\makeatletter
  \newcommand\figcaption{\def\@captype{figure}\caption}
  \newcommand\tabcaption{\def\@captype{table}\caption}
\makeatother

\title{\LARGE \bf
COMPASS: Contrastive Multimodal Pretraining \\for Autonomous Systems
}

\author{Shuang Ma$^{\dagger}$, Sai Vemprala$^{\dagger}$,  Wenshan Wang$^{\star}$, Jayesh K. Gupta$^{\dagger}$, \\
Yale Song$^{\dagger}$, Daniel McDuff$^{\dagger}$ and Ashish Kapoor$^{\dagger}$
\thanks{$^{\dagger}$ Microsoft Redmond, WA
        {\tt\small \{shuama,sai.vemprala, jayesh.gupta,yalesong,damcduff,akapoor\} @microsoft.com}}%
\thanks{$^{\star}$ Carnegie Mellon University Pittsburgh, PA
        {\tt\small wenshanwang@cmu.edu}}%
}

\begin{document}

\maketitle
\thispagestyle{empty}
\pagestyle{empty}

\begin{abstract}
Learning representations that generalize across tasks and domains is challenging yet necessary for autonomous systems. Although task-driven approaches are appealing, designing models specific to each application can be difficult in the face of limited data, especially when dealing with highly variable multimodal input spaces arising from different tasks in different environments.We introduce the first general-purpose pretraining pipeline, COntrastive Multimodal Pretraining for AutonomouS Systems (COMPASS), to overcome the limitations of task-specific models and existing pretraining approaches. COMPASS constructs a multimodal graph by considering the essential information for autonomous systems and the properties of different modalities. Through this graph, multimodal signals are connected and mapped into two factorized spatio-temporal latent spaces: a ``motion pattern space'' and a ``current state space.'' By learning from multimodal correspondences in each latent space, COMPASS creates state representations that models necessary information such as temporal dynamics, geometry, and semantics. We pretrain COMPASS on a large-scale multimodal simulation dataset TartanAir~\cite{tartanair2020iros} and evaluate it on drone navigation, vehicle racing, and visual odometry tasks. The experiments indicate that COMPASS can tackle all three scenarios and can also generalize to unseen environments and real-world data.\footnote{Our code implementation can be found at \url{https://github.com/microsoft/COMPASS}}.


\end{abstract}

\section{Introduction}
\label{sec:intro}

A fundamental facet of human intelligence is the ability to perceive the environment and encode multimodal sensory signals into complex neural representations \cite{Andersen1997,lacey2016crossmodal}, which are then used to complete a wide variety of tasks. Similarly, learning representations that capture the underlying state of an environment from different sensors, while taking into account an agent's dynamic capabilities is crucial for autonomous systems. Such concise, jointly learned representations have the potential to effectively transfer knowledge across tasks and enable learning with fewer environmental interactions. The ability to perceive and act is crucial for any embodied autonomous agent and is required in many situations involving different form factors and scenarios. For example, localization (or being able to answer ``Where am I?'') is a fundamental question that needs to be answered by any autonomous agent prior to navigation, this is often achieved via visual odometry. Highly dynamic tasks, such as vehicle racing, necessitate collision avoidance and require precise understanding for planning a trajectory and meeting objectives. In both cases learning geometric and semantic information from the environment is crucial. Task-specific approaches produce promising results, but they involve learning only the part of information tailored for the intended tasks, which can be limiting in utility by failing to generalize to new scenarios. 
We investigate whether it is possible to build a general-purpose pretrained models in a task-agnostic fashion, which can be useful in solving various downstream tasks relevant to the perception-action loops in autonomous systems. 

Although pretrained models have shown strong performance in domains such as NLP~\cite{devlin2018bert, brown2020GPT-3} and computer vision~\cite{chen2020SimCLR,grill2020bootstrap}, building such models for autonomous systems brings unique challenges. 
First, the environments are usually perceived through multimodal sensors, so the model needs the ability to make sense of multimodal data. 
Existing multimodal learning approaches primarily focus on mapping multimodal data into joint latent spaces \cite{shuang2019:multimodalIB, sun2019learning, tian2019CMC}. These approaches are suboptimal for autonomous systems as they do not address aspects such as differing sampling rate, temporal dynamics, and geo-centric or object-centric spatial factors. These are crucial factors in our scenario due to variations that arise from sensor and actuator configurations in autonomous systems. Secondly, autonomous systems deal with a complex interplay between perception and action. The target learning space is highly variable due to a large variety of environmental factors, application scenarios, and system dynamics. This is in stark contrast to language models that focus on underlying linguistic representations, or visual models centered on object semantics. Finally, unlike NLP and computer vision, there is a scarcity of multimodal data that can be used to train large pretrained representations for autonomous systems.

In this work, we introduce COntrastive Multimodal Pretraining for AutonomouS Systems (COMPASS), a multimodal pretraining approach for {\em perception-action loops}. 
COMPASS builds a general-purpose representation that generalizes to different environments and tasks. Unlike the prevalent approaches, COMPASS aims to learn a generic representation by exploiting underlying properties across multiple modalities, while appropriately considering the dynamics of the autonomous system. Self-supervised learning using a  large corpus of multimodal data collected from various environments allows the model to be completely agnostic to downstream tasks. 

\begin{figure}
    \centering
    \includegraphics[width=0.5\textwidth]{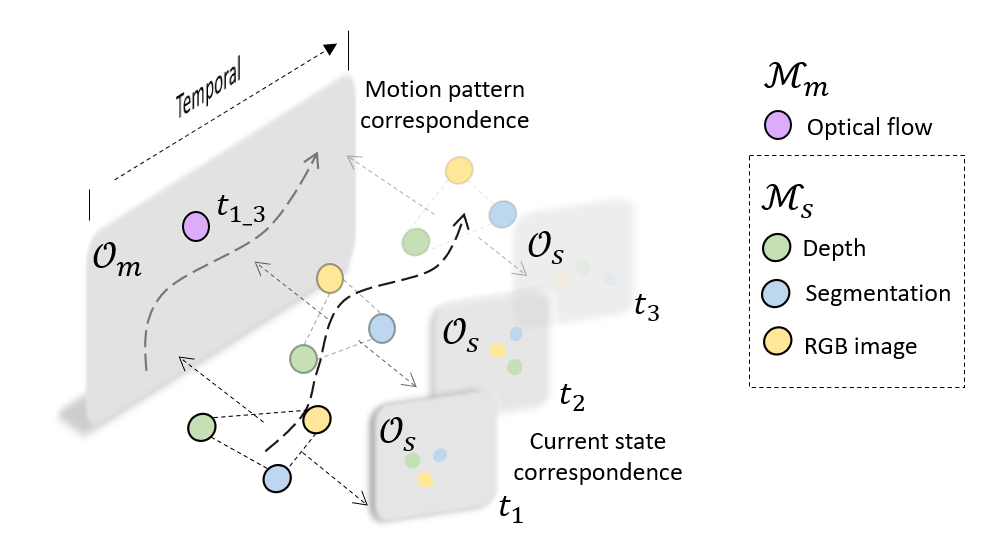}
    \caption{We introduce COntrastive Multimodal Pretraining for AutonomouS Systems (COMPASS). Given multimodal signals of spatial and temporal modalities $\mathcal{M}_{s}$ and $\mathcal{M}_{m}$, respectively. COMPASS learns two factorized latent spaces, i.e., a motion pattern space $\mathcal{O}_m$ and a current state space $\mathcal{O}_s$, using multimodal correspondence as the self-supervisory signal.}
    \label{fig:graph-multimodal}
\end{figure}

Our design choices are informed by seeking answers to two questions: 
1) \textit{What information would be essential to solve common tasks in autonomous systems?}
2) \textit{How can we represent such information by learning from multisensory multimodal data captured by autonomous agents?} 
First, we posit that information essential for autonomous systems lies in a spatio-temporal space that models motion (ego-motion or environmental), geometry and semantic cues. We also observe that such information is typically perceived by an autonomous agent through multimodal sensors. Consequently, we propose a multimodal graph as a core building block that models such spatio-temporal relationships and statistical characteristics of different modalities (\Cref{fig:graph-multimodal}). Intuitively, the graph maps all modalities into a factorized spatio-temporal latent space comprising of two subspaces: a \emph{motion pattern space} and a \emph{current state space}. The first subspace explicitly models and handles the temporal and system dynamics of autonomous systems, while the latter is designed to encode geometric and semantic information coming from modalities representing the states at certain local time points, e.g. a single RGB frame. Training COMPASS is then geared towards learning to associate multimodal data from a large training corpus. Such a factorized representation captures important spatio-temporal structure important for various downstream tasks while allowing different sensors to use the same pretrained model. By evaluating the pretrained COMPASS on three downstream tasks, i.e. {\em Vehicle Racing}, {\em Visual Odometry}, and {\em Drone Navigation}, with variations across environments, dynamics, and application scenarios, we observe that COMPASS generalizes well to different tasks, unseen environments and real-world challenges even in the low-data regimes.


\section{Related Work}
Representation learning has been an area of great interest in machine learning as well as in robotics. Self-supervised learning has been shown to be effective in vision particularly through the use of contrastive objectives~\cite{oord2018CPC, wang2019improved, chen2020SimCLR, grill2020bootstrap}. Recently, there is growing interest in learning ``object-centric'' representations of visual scenes \cite{bear2020learning, locatello2020object}. Contrastive learning has also been applied to reinforcement learning to match data augmentations with raw observations \cite{srinivas2020curl}. 

Learning multimodal representations has been examined in several domains such as vision-language \cite{li2020unicoder,ahuja2020no}, vision-audio \cite{ginosar2019learning,owens2018audio}, image registration \cite{roche1998correlation,hu2018weakly}, and video understanding \cite{sun2019learning, gordon2020watching}. Tsai et al.~\cite{tsai2018learning} present a framework learning intra-modal and cross-modal interactions from input, and Alayrac et al.~\cite{alayrac2020self} present a multimodal learning approach for text, audio and video. 
Inspired by the success of large-scale pretraining in the text domain \cite{devlin2018bert, brown2020GPT-3}, pretrained models have also been developed for vision-language tasks \cite{lu2019vilbert, zhang2021vinvl, li2020oscar}. A natural extension of multimodal learning algorithms has been applied to the multi-task learning setting \cite{pramanik2019omninet, kuga2017multi, chaplot2019embodied}. Numerous surveys on multimodal learning are also largely focused on vision, text and speech~\cite{ramachandram2017deep,zhang2020multimodal,guo2019deep}. Baltrusaitis et al.~\cite{baltruvsaitis2018multimodal} point out the opportunity for co-learning with multimodal data where knowledge from one (resource rich) modality can be exploited in modeling another (resource poor) modality.  

Autonomous systems require rich, well-grounded representations and benefit from the existence of multiple sensors of different modalities. Robotics tasks such as manipulation have been shown to benefit from object-centric representations \cite{devin2018deep, florence2018dense}, and combining geometry has been shown to be effective for navigation tasks. Multimodal representation learning has been applied to robotic manipulation and grasping in the form of visuo-tactile representations \cite{lee2019making}, as well as scene understanding and exploration by combining RGB and depth \cite{chaplot2020object}, and human robot interaction \cite{campbell2019probabilistic}. Cross-modal representation learning has been combined with imitation learning to result in drone navigation policies \cite{bonatti:2020:cmvae}. Multi-task learning has been examined for self-driving under different modes \cite{chowdhuri2019multinet} and visual odometry/semantic segmentation \cite{radwan2018vlocnet++}. 
Instead of leveraging specific designs which are tailored for each of these tasks, here we propose a general-purpose representation learning approach in the context of the perception stack of an embodied autonomous agent. With the single unified pretraining model, different tasks can hence be easily achieved with a fast finetuning on a small amount of data.

\section{Approach}
We set out to create a model that can be pretrained on simulated environments providing data from different modalities. For example, each environment may provide any combination of RGB frames, depth, and optical flow. Our goal is to learn a model that can ingest any of these modalities and produce \emph{general-purpose} representations that can be further finetuned for various autonomous tasks.
We posit that autonomous systems need representations that encode both the current state of the environment as well as the temporal dynamics.
As such, we propose a \textit{multimodal graph} that respects the underlying spatio-temporal relationship between modalities (\Cref{sec:review}), and use it as the basis for designing contrastive learning objectives (\Cref{sec:training-obj}) to learn two factorized latent spaces that model them, respectively.

\subsection{Multimodal Graph Construction}
\label{sec:review}

Given a set of multimodal data $\{\mathcal{M}\}_N$ with $N$ modalities, there are several ways to learn representations from it. We will first briefly review the current multimodal learning approaches, and then introduce our spatio-temporal multimodal graph.

\textbf{Joint common multimodal latent space.} The most straightforward way is to simply map all modalities into a common latent space. Thus, different modalities can model their common underlying information through the joint space. For example in \cite{shuang2019:multimodalIB}, visual, audio and text signals are associated based on the semantics. These approaches have the advantage of simple design, but they are based on an assumption that all modalities share a single underlying factor, say semantic information, and hence can be equally treated, for example, mapping through a single projection head. 

This approach suffers from the issue that different complementary properties across various modalities are not fully utilized. Projection to a shared single latent space results in a loss of information: for example, na\"ively projecting an optical flow map with an depth map into a single space is not ideal as they encode temporal dynamics and geometry respectively, which are very different, and it is important to maintain that distinction while learning from such modalities.

\textbf{Disjoint cross-modal latent space.} Another line of work learns multimodal data by capturing the cross-modal relation between each pair of modalities. For example, Tian et al.~\cite{tian2019CMC} extend cross-modal contrastive objective to a multimodal/multi-view setting. Connections are set by contrasting each modality pairs in disjoint latent spaces. While such a strategy enables the specificity of different modality pairs, scaling it to a more complex setting, e.g., more than two modalities, presents several computational challenges. 

\begin{figure}
\centering
  \includegraphics[width=0.5\textwidth]{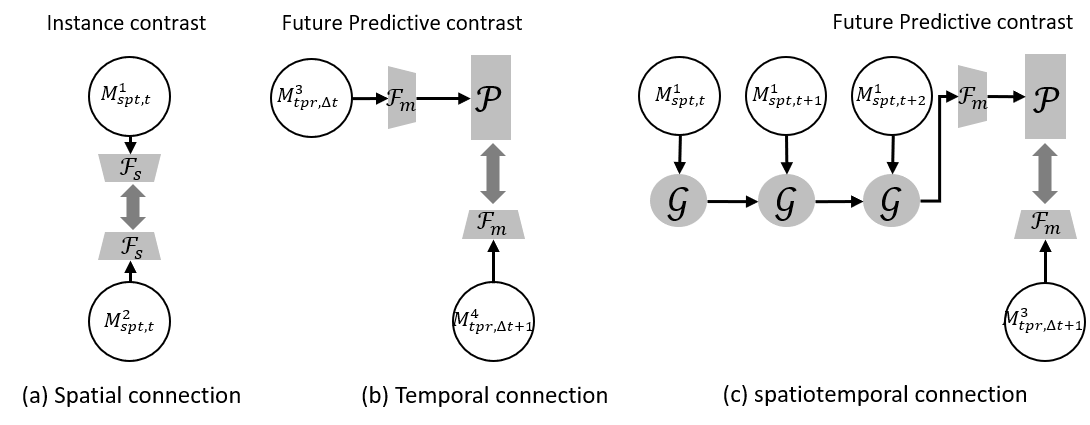}
  \caption{Three types of connections. $\mathcal{M}_{s,t}^i$: $i$-th spatial modality at time step $t$. $\mathcal{M}_{m, \delta t}^i$: $i$-th temporal modality in a time window $\delta t$. $z_t^i$: extracted representation from modality $i$ at time $t$, $z'^{i}_t$: the latent code mapped by spatial projection head $\mathcal{F}_s$. $z'^i_{t \to t+j}$: the latent code projected by temporal projection head $\mathcal{F}_m$. $c_{t \to t+j}^i$: context vector induced by aggregation network $\mathcal{G}$. $\hat{z}$: predicted latent code. $E_i$: modality encoder for modality $i$. Different color signifies different modality. The modules in shaded gray share weights among all modalities, i.e. $\mathcal{F}_s$, $\mathcal{F}_m$, $\mathcal{G}$ and $\mathcal{P}$.}
  \label{fig:connection-type}
\end{figure}

\textbf{Spatio-temporal Multimodal latent space.} The model we propose enables us to model the complementary information across the multimodal signals in a scalable manner via a novel multimodal graph design. The key insight we build upon is that the information essential for autonomous systems lies in a spatio-temporal space that can be partitioned to model temporal statistics and spatial and/or semantic aspects. Consequently, we start by first categorizing our multimodal data streams into two classes: \textit{spatial} modalities and \textit{temporal} modalities. Given a set of $N$ multimodal data streams $\{\mathcal{M}\}_N$ with $n$ spatial modalities $\{\mathcal{M}_{s}\}_{n}$ and $l$ temporal modalities $\{\mathcal{M}_{m}\}_l$, we then jointly learn two latent spaces, a ``motion pattern space'' $\mathcal{O}_{m}$ and a ``current state space'' $\mathcal{O}_{s}$. 

Let $ \{\mathcal{M}_{m}^1, \mathcal{M}_{m}^2, ..., \mathcal{M}_{m}^{l} \}_{\delta T}$ denote a data sequence from temporal modalities that are time-synchronized within a window $\delta T$, which share a certain motion pattern arising from the agent or the environment. As shown in \Cref{fig:connection-type}, we construct temporal connections by mapping them to the common motion pattern space $\mathcal{O}_m$ through a projection head $\mathcal{F}_m$, i.e. $\mathcal{F}_m({\{\mathcal{M}_{m}^i\}_{\delta T}^{i\in[1,l]}}) \to \mathcal{O}_m$. Similarly, we construct spatial connections by mapping data from spatial modalities at each discrete time step $t$, i.e., $\{\mathcal{M}_{s}^1, \mathcal{M}_{s}^2, ..., \mathcal{M}_{s}^n \}_{t}$, to the common current state space $\mathcal{O}_s$ through a projection head $\mathcal{F}_s$, i.e. $\mathcal{F}_s(\{{\mathcal{M}_{s}^i\}_t^{i\in[1,n]}}) \to \mathcal{O}_s$. Furthermore, to better associate spatial modalities with the temporal ones, we rely on the intuition that multiple instances of one/more spatial modalities from a window of time can be associated with observations from a temporal modality from that same window of time. Thus, a spatio-temporal connection is added by aggregating sequential data from spatial modalities using an aggregation head $\mathcal{G}$ and mapping them to the common motion pattern space $\mathcal{O}_m$ using the projection head $\mathcal{F}_m$, i.e. $\mathcal{F}_m(\mathcal{G}(\{\mathcal{M}_{s, t}^i, \mathcal{M}_{s, t+1}^i, ..., \mathcal{M}_{s, t+\delta t}^i \}^{i\in[1,n]})) \to \mathcal{O}_m$.


\begin{figure*}
    \centering
    \includegraphics[width=.7\textwidth]{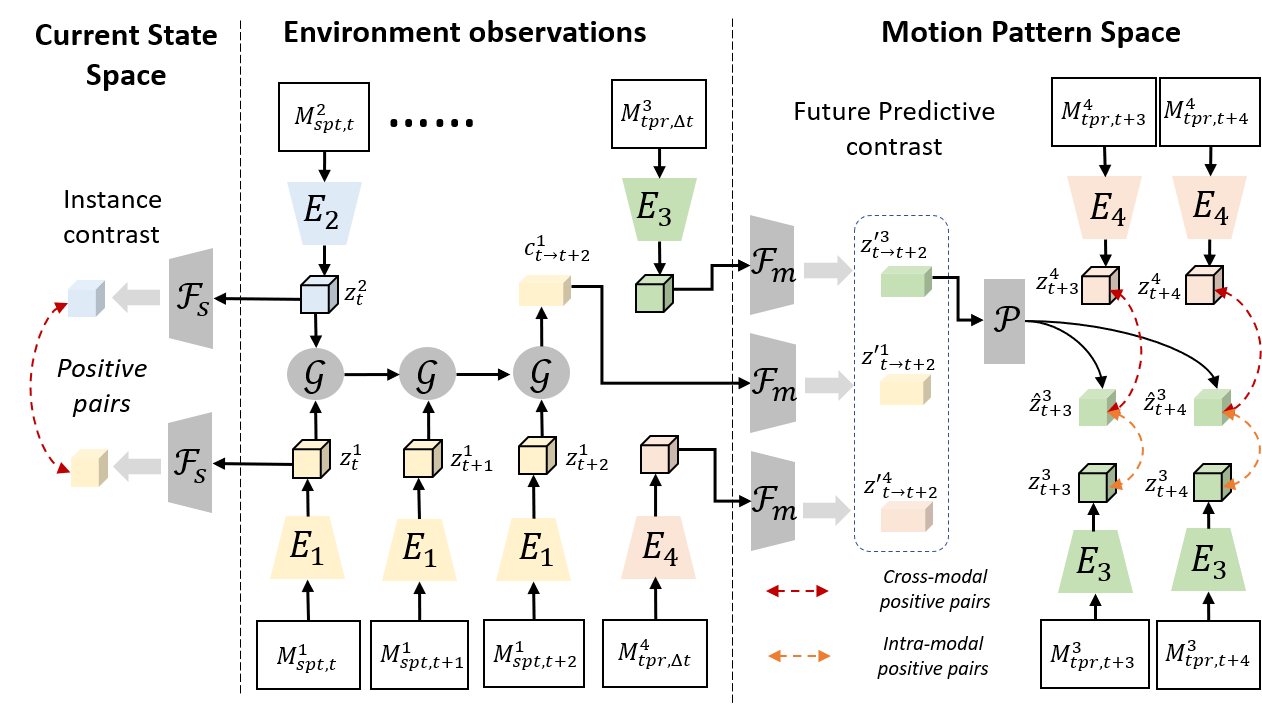}
    \caption{Pretraining pipeline of COMPASS. $\mathcal{M}_{s,t}^i$ denotes the $i$-th spatial modality at time step $t$. $\mathcal{M}_{m, \delta t}^i$ denotes the $i$-th temporal modality in a time window $\delta t$. $z_t^i$ denotes the extracted representation from modality $i$ at time $t$, $z'^{i}_t$ is the latent code mapped by spatial projection head $\mathcal{F}_s$. $z'^i_{t \to t+j}$ is the latent code projected by temporal projection head $\mathcal{F}_m$. $c_{t \to t+j}^i$ denotes the context vector induced by aggregation network $\mathcal{G}$. $\hat{z}$ represents the predicted latent code. $E_i$ is the modality encoder for modality $i$. Different color signifies different modality. The modules in shaded gray share weights among all modalities, i.e. $\mathcal{F}_s$, $\mathcal{F}_m$, $\mathcal{G}$ and $\mathcal{P}$.}
    \label{fig:pipeline}
\end{figure*}

The multimodal graph with spatial, temporal, and spatio-temporal connections serves as a framework for learning multimodal representations by encoding the underlying properties of modalities (such as modality specific state information, dynamics) as well as any common information shared between them (for example global geometry and motion). These connections enable information sharing across all the modalities through the defined spatio-temporal latent spaces. Next we discuss the contrastive objectives to learn the parameters of the model that includes the factorized latent spaces $\mathcal{O}_{m}$ and $\mathcal{O}_{s}$s.

\subsection{Training Objective} \label{sec:training-obj}

\textbf{Contrastive Objective for Temporal Connections.}
To encode temporal information in the motion pattern space $\mathcal{O}_m$, we solve a contrastive learning objective that associates pairs of time-synced data from different modalities. Intuitively, if a model successfully captures temporal information from one modality, it should have the predictive capacity to model a few future time steps for itself as well as the other modalities. We formulate this intuition into a contrastive learning objective.

Given a set of $m$ time-synced temporal modalities, we define positive pairs as a sequence of embeddings for the true observations $[z_{t+1}, z_{t+2}, ..., z_{end}]_{\mathcal{M}^{i\in[1, m]}}$ and a sequence of embeddings predicted recursively from an anchor modality $[\hat{z}_{t+1}, \hat{z}_{t+2}, ... , \hat{z}_{end}]_{\mathcal{M}^a}$, where $\mathcal{M}^a$ is the anchor modality. The positive pairs from $m$ modalities include the true future observations of anchor modality, its own, and the remaining $m-1$ modalities. Thus, the comparison is performed both in an intra-modal and a cross-modal fashion.
As shown in \Cref{fig:pipeline}, the modality-specific encoders $E$, extract embeddings from each modality. These are then mapped to the common motion pattern space $\mathcal{O}_{m}$ through the motion pattern projection head $\mathcal{F}_m$. A prediction head $\mathcal{P}$ is added on top to perform future prediction. The contrastive loss is computed between the predicted future representations and their corresponding encoded true representations. 
Our contrastive objective is then:
\begin{equation}
    \mathcal{L}_{m} = -\sum_{t, i} \log \frac{\exp(\hat{z}_{t, a}^T z_{t,i})}{\exp(\hat{z}_{t, a}^T z_{t,i})+\sum_{j\neq t} \exp(\hat{z}_{t,a}^T z_{j, i})}
    \label{eq:temporal-loss}
\end{equation} 
where $t, i, a$ denote the time, modality, and anchor indices.

\textbf{Contrastive Objective for Spatial Connections.}
The current state space $\mathcal{O}_s$ is expected to encode geometric and semantic information by associating the data from spatial modalities $\mathcal{M}_{s}$ together. We again utilize a contrastive objective to formulate this idea. 

Given $n$ spatial modalities, we define the positive pairs as the representation from an anchor modality at time step $t$, i.e. $z_{t, a}$, and the representations induced by all the spatial modalities at the same time step $t$, i.e. $\{z_{t, i}\}_{i\in[1,n]}$. The negative pairs are sampled from representations induced by spatial modalities at different time steps. We formulate this instance level contrastive loss as:
\begin{equation}
    \mathcal{L}_{s} = -\sum_{t, i} \log \frac{\exp(z_{t,a}^T z_{t,i})}{\exp(z_{t,a}^T z_{t,i})+\sum_{j\neq t} \exp(z_{t,a}^T z_{j, i})}
    \label{eq:spatial-loss}
\end{equation}
where $t, i, a$ denote time, modality and anchor, respectively. Note that the anchor representation is sampled from $\mathcal{O}_s$, i.e. $\mathcal{F}_s( E(\mathcal{M}_a)) \to z_{a}$. This is different from \Cref{eq:temporal-loss} where the anchor is the estimated representation induced by the prediction head, i.e. $\mathcal{P}(\mathcal{F}_m(E(\mathcal{M}_a))) \to \hat{z}_a$. Therefore, while \Cref{eq:temporal-loss} computes contrastive loss through future prediction, \Cref{eq:spatial-loss} computes it at an instance level.


\textbf{Objective for Spatio-temporal Connections.}
The spatio-temporal connections encode motion patterns from consecutive observations of spatial modalities. Given a sequence $[M_{a,t}, M_{a, t+1}, ..., M_{a, t+\delta t}]$ from an anchor modality $\mathcal{M}_a \in \{\mathcal{M}_{s}\}_n$, we obtain embeddings using the modality encoder, i.e. $E([M_{a,t}, M_{a, t+1}, ..., M_{a, t+\delta t}]) \to [z_{a,t}, z_{a, t+1}, ..., z_{a, t+\delta t}]$. We then use the aggregation network $\mathcal{G}$ to project them to $\mathcal{O}_m$ and produce an aggregated context vector $c_a$, i.e. $\mathcal{G}([z_{a,t}, z_{a, t+1}, ..., z_{a, t+\delta t}]) \to c_a$. Given this context vector, we can compute future predictions similar to the way the motion pattern loss was computed, i.e., by inputting $c_a$ to $\mathcal{P}$ for future prediction as $\mathcal{P}(c_a) \to \hat{z}_a$.
To this end, we again utilize \Cref{eq:temporal-loss} to minimize the contrastive objective of $\mathcal{L}_{sm}$. Our learning objective is: $\mathcal{L}=\mathcal{L}_m + \mathcal{L}_s + \mathcal{L}_{sm}$. 

After pretraining, the COMPASS model can be finetuned for several downstream tasks. Based on the sensor modalities available for the task of interest, we connect the appropriate pretrained COMPASS encoders to prediction heads responsible for task-specific predictions. This combined model is then finetuned given data and objectives from the specific task. 
\section{Experiments}

The experiments aim to demonstrate the effectiveness of COMPASS as a general-purpose pretraining approach. We tackle three downstream scenarios that are representative of autonomous system tasks: vehicle racing (\Cref{sec:car-racing}), visual odometry (\Cref{sec:visual-odometry}), and drone navigation (\Cref{sec:drone-navigation}), for all of which we finetune a single pretrained COMPASS model. 

Through our experiments, we explore the following questions:\\ 
\textit{1) Can COMPASS adapt to unseen environments and real-world scenarios?}
COMPASS is pretrained on simulation data (TartanAIR \cite{tartanair2020iros}) and we demonstrate experiments on a real-world benchmark (KITTI \cite{Geiger2013KITTI}) to understand sim2real performance (\Cref{sec:visual-odometry}). Similarly, our experiments with the vehicle racing task investigate how well we can generalize to completely unseen environments (\Cref{sec:car-racing}). \\
\textit{2) What are the benefits of COMPASS when compared to other representation learning approaches?}
We compare COMPASS with task-specific approaches and representative pretraining/multimodal learning approaches in \Cref{sec:car-racing}). \\
\textit{3) Can COMPASS improve data efficiency?}
We compare models finetuned over COMPASS representations to task specific models trained from scratch as we vary the data set size, as we analyze the learning performance (\cref{sec:drone-navigation}).


\begin{figure}
    \centering
    \includegraphics[width=\columnwidth]{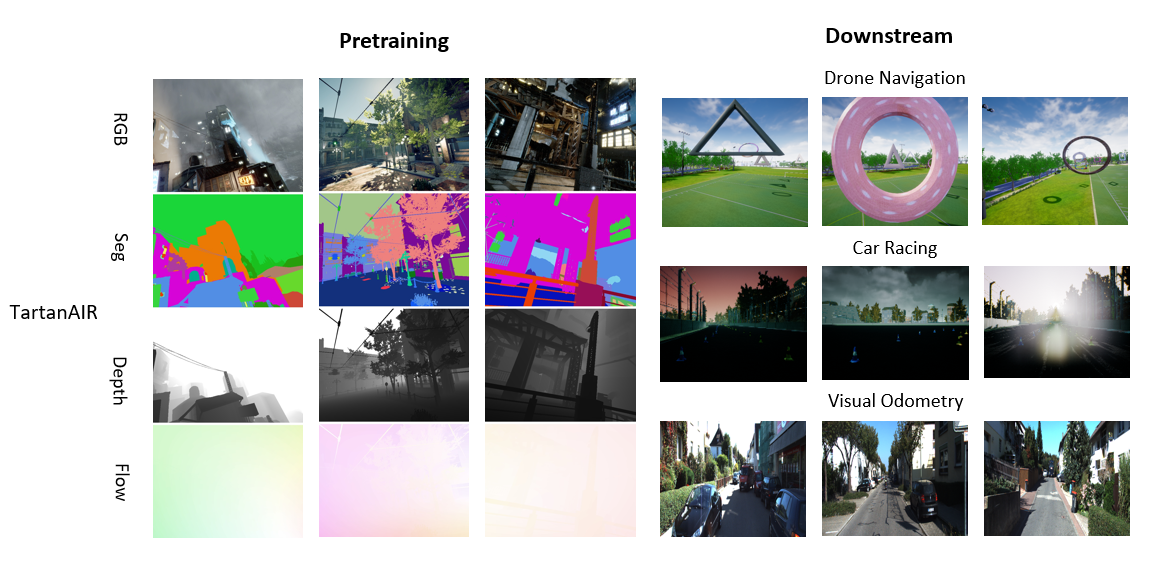} 
    \caption{Samples from TartanAIR and the downstream task datasets. Notice the difference in the visual scene: a soccer field in daylight (drone navigation), a racing track with varying backgrounds (vehicle-racing), and a real world scene (VO).}
    \label{fig:data-sample}
\end{figure}

\begin{table}[]
    \centering
    \caption{Various datasets used in our experiments.}
    \begin{tabular}{llrr}
    \toprule
         Dataset       &  Usage                  &  Scale    & Env. \\ \midrule
         TartanAIR  &  Pretrain              & 1M         & 16    \\
         AirSim-Car    &  Vehicle Racing   & 17k      & 9     \\ 
         KITTI         &  Visual Odometry & 23K      &  11 \\
         Drone-Gate   &  Drone Navigation         & 3k     & 1 \\ \bottomrule
    \end{tabular}
    \label{tab:data-summery}
\end{table}

\textbf{Pretraining.} We use a 3D-ResNet18~\cite{hara2018can} architecture for the encoder for each modality $E_m$, a two-layer CNN for the future prediction head $P$, and a bidirectional ConvGRU~\cite{ballas2015delving} for the aggregation head $G$ (shared across modalities) We use the TartanAIR~\cite{tartanair2020iros} dataset for pretraining that contains 400K sensor samples from diverse environments including indoor, outdoor, urban, nature, and sci-fi scenes. The dataset is generated with a simulated pinhole camera, and provides multimodal signals. We pretrain COMPASS on 16 environments of TartanAIR with data from three modalities: RGB, depth, and optical flow. Sample data from the pretraining dataset can be seen in \Cref{fig:data-sample} along with the downstream task datasets. In \Cref{tab:data-summery}, we list some details about the extent of data used for pretraining and task-specific finetuning. 

\subsection{Vehicle Racing} \label{sec:car-racing}
\textbf{Task and setting.} The goal here is to enable autonomous vehicles to drive in a competitive Formula racing environment. We use the AirSim-Car dataset~\cite{zadok2019car-racing} that provides 9 simulated racetrack environments in the AirSim simulator, each with 2 lanes separated with different colored traffic cones. The environment contains visual distractors such as ad signs, tires, grandstands, and fences, which help add realism and increase task difficulty. The control module is expected to predict the steering angle such that the car can successfully maneuver through the tracks and avoid obstacles.
We construct a perception module with the RGB encoder from \texttt{COMPASS} pretrained on TartanAIR and define a control module as a two-layer MLP with a prediction head that outputs the steering wheel angle (normalized to $[0, 1]$). We finetune the model on the AirSim-Car dataset with ${L}1$ loss measuring the per-frame angle discrepancy. 

\textbf{Baselines.} We compare COMPASS with a model trained from scratch (\texttt{Scratch}), 2 pretraining approaches (\texttt{CPC}~\cite{oord2018CPC} and \texttt{CMC}~\cite{tian2019CMC}), and 2 multimodal learning approaches (\texttt{JOINT} and \texttt{DISJOINT}). \texttt{Scratch} is directly trained on the AirSim-Car dataset without pretraining, whereas the pretraining and multimodal learning approaches are pretrained on TartanAIR before finetuning on the AirSim-Car dataset. More details on the baselines:

\begin{itemize}[noitemsep,leftmargin=*]
    \setlist{nolistsep}
    \item \texttt{Scratch} trains a randomly initialized network (the same architecture as ours) from scratch.
    \item \texttt{CPC}~\cite{oord2018CPC} is a contrastive learning approach that learns representations by predicting the future representations in the latent space.
    \item \texttt{CMC}~\cite{tian2019CMC} is a contrastive learning approach that captures information shared across modalities. Unlike \texttt{CPC}, it learns from multiple views and the contrastive loss is defined at an instance level rather than in a predictive manner. 
    \item \texttt{JOINT} learns multimodal data from a single joint latent space by mapping modalities with a single projection head.
    \item \texttt{DISJOINT} learns multimodal data from disjoint latent spaces. Other than using a single projection head, it creates a cross-modal latent space for each modality pairs, and all of the latent spaces are disjoint. 
\end{itemize}

\begin{table}[]
    \centering
        \caption{Results on vehicle racing experiments. The numbers are mean/std of L1 errors on steering angle prediction over 5 runs.}
    \begin{tabular}{lcc}
    \toprule
        Model    & Seen environment & Unseen environment   \\ \midrule
        \textsc{Scratch}  & 0.085 $\pm$ 0.019     & 0.120 $\pm$ 0.009 \\
         \textsc{CPC}      & \textbf{0.037} $\pm$ 0.012    & 0.101 $\pm$ 0.017 \\
        \textsc{CMC}      & 0.039 $\pm$ 0.013    & 0.102 $\pm$ 0.012 \\
        \textsc{Joint}    & 0.055 $\pm$ 0.016    & 0.388 $\pm$ 0.018 \\
        \textsc{Disjoint} & 0.039  $\pm$ 0.017    & 0.131 $\pm$ 0.016 \\ 
        \textsc{COMPASS}  & 0.041 $\pm$ 0.021     & \textbf{0.071} $\pm$ 0.023\\
        \bottomrule
    \end{tabular}
    \label{tab:car-race-compare}
\end{table}


\textbf{Can COMPASS generalize to unseen environments?} We explore the hypothesis pretraining can help with generalization to unseen environments. Consequently, we compare \texttt{COMPASS} with \texttt{Scratch} (no pretraining) and the other pretraining approaches: \texttt{CPC}, \texttt{CMC}, \texttt{JOINT}, and \texttt{DISJOINT}. We evaluate these models in two settings: 1) trained and evaluated on the same environments (``seen''); 2) trained and evaluated on different environments (``unseen''). \Cref{tab:car-race-compare} shows that overall, there is a large gap between \texttt{Scratch} and the other pretraining approaches. The performance degradation in the unseen environment is relatively marginal with \texttt{COMPASS}, which suggests its effectiveness compared to the other pretraining approaches. 

\begin{figure}[!t]
        \centering
        \begin{subfigure}[t]{0.47\columnwidth}
            \centering            
            \includegraphics[width=\textwidth]{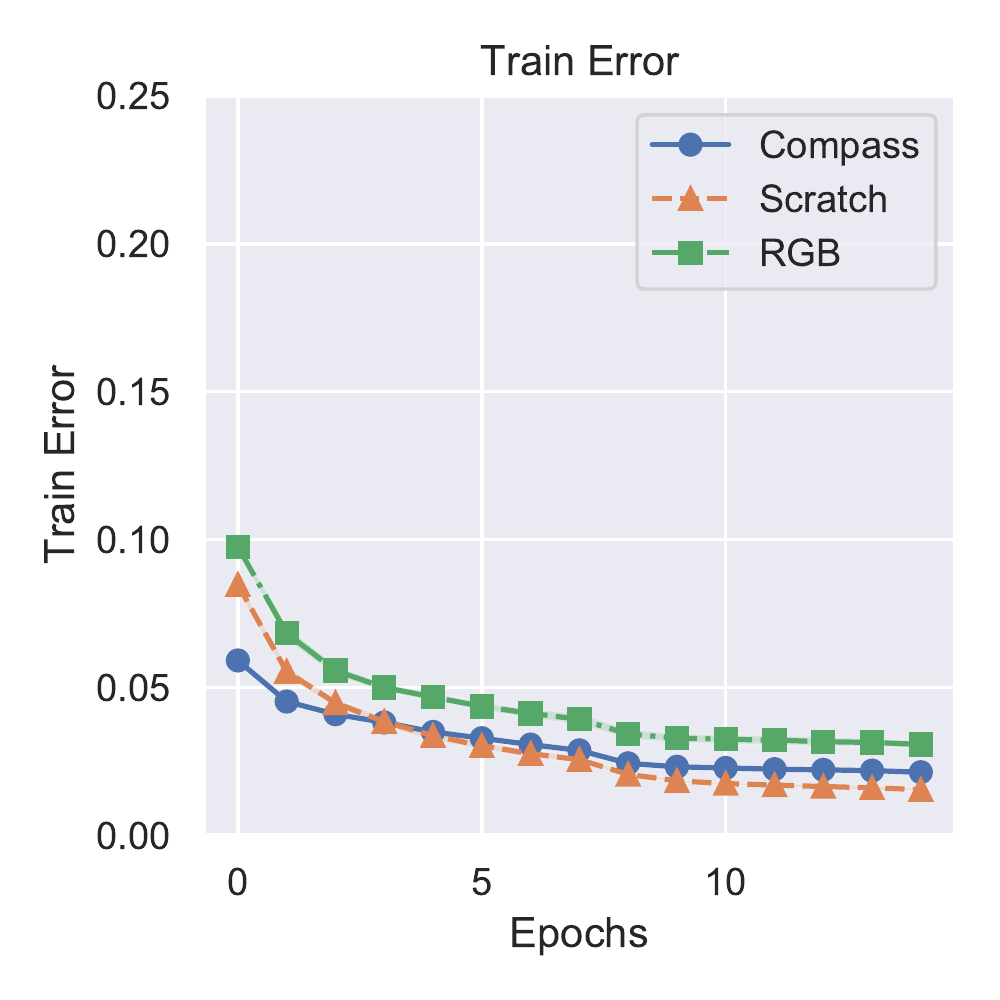}
            \caption{Train error profile}
            \label{fig:car_steering_train}
        \end{subfigure}
        \begin{subfigure}[t]{0.47\columnwidth}
            \centering            
            \includegraphics[width=\textwidth]{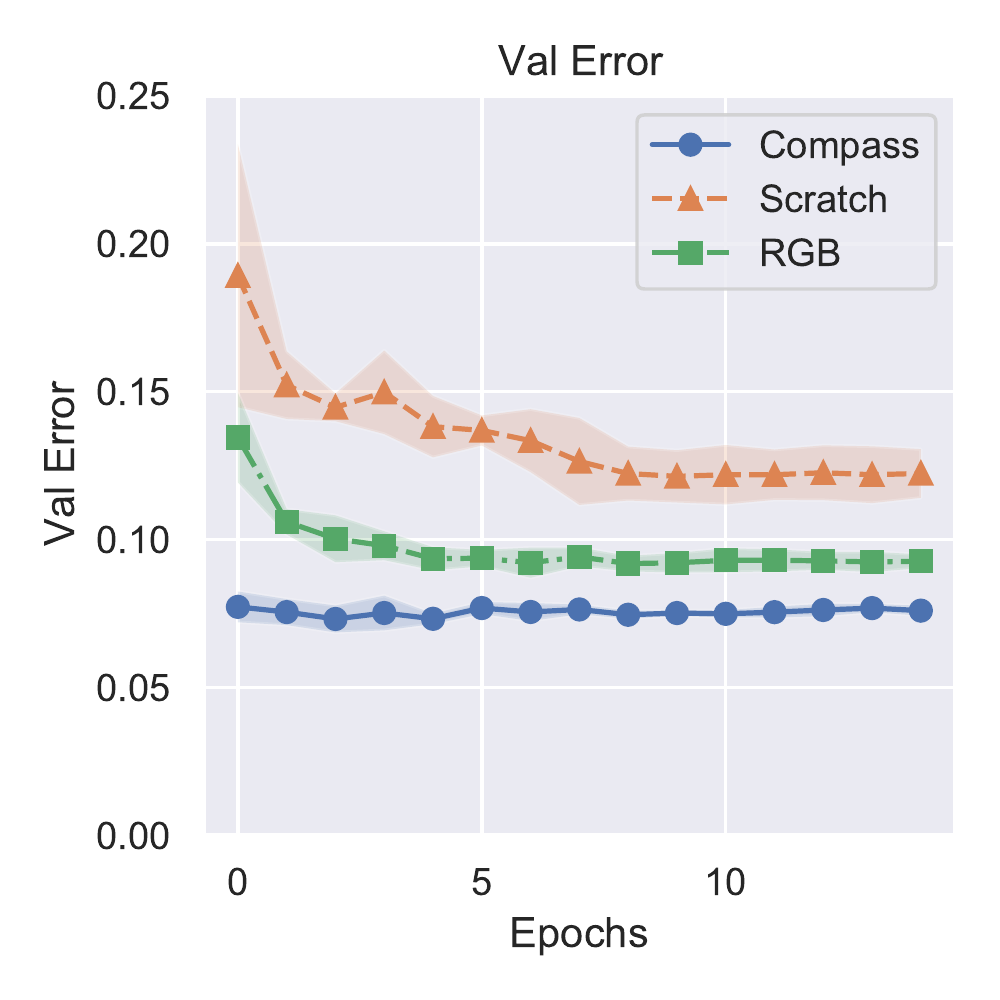}
            \caption{Validation error profile}
            \label{fig:car_steering_val}
        \end{subfigure}
    \caption{Comparison of train and test error profiles for vehicle racing between COMPASS, a model pretrained only with RGB modality, and a model trained from scratch for the task.}
    \label{fig:car_steering_error}
\end{figure}

\textbf{Can COMPASS benefit from multimodal pretraining regime?} 
We investigate the effectiveness of pretraining on multimodal data by analyzing loss curves from different pretrained models on the same ‘unseen’ environments. \Cref{fig:car_steering_error} compares the train/validation loss curves of \texttt{COMPASS}, \texttt{RGB}, and \texttt{Scratch}. Here \texttt{RGB} is the model pretrained by using the same backbone and training objectives with \texttt{COMPASS}, but only pretrained with RGB modality. By comparing \texttt{COMPASS} with \texttt{RGB}, we observe that pretraining on multimodal data helps COMPASS achieve the best performance overall. Also, both of these pretraining models show large gaps when compared to a model trained from scratch (\texttt{Scratch}). When comparing \Cref{fig:car_steering_train} to \Cref{fig:car_steering_val}, we also see that \texttt{Scratch} suffers more from an overfitting issue than the other two models.


\subsection{Visual Odometry} \label{sec:visual-odometry}

\textbf{Task and setting.} 
Visual odometry (VO) aims to estimate camera motion from consecutive image frames. It is a fundamental component in visual SLAM and is a widely used for localization in robotics. Note that we focus on a simple visual odometry task, which only takes in two consecutive images as inputs. This is different from full-fledged SLAM systems~\cite{mur2015orb, yang2018deep,yang2020d3vo}, which utilize key-frame optimization in the back-end - our task can be considered as the pose tracking module in the SLAM front-end.

We evaluate COMPASS for the VO task using a real-world dataset KITTI~\cite{Geiger2013KITTI} which is one of the most widely used benchmarks in the VO/SLAM literature. It contains 11 labeled sequences including 23,201 image frames in a driving scenario. On this dataset, we examine the generalization ability of COMPASS that was pretrained purely on simulation data to the real-world data. 
We attempt the VO task using a model designed as a two-stage structure, following TartanVO~\cite{tartanvo2020corl}. In the first stage, a pretrained optical flow estimation network, PWC-Net~\cite{sun2018pwc}, is utilized to convert consecutive RGB image pairs into optical flow images by extracting the dense correspondence information. In the second stage, a pose network estimates the camera motion based on the optical flow. In our case, we utilize the pretrained optical flow encoder from COMPASS  coupled with a predictin head as the second-stage pose network, so that we can evaluate the effectiveness of COMPASS for the flow modality. The model is asked to estimate the camera translation and rotation.

\begin{table}[]
    \centering
    \caption{Comparison of translation and rotation errors on KITTI dataset. VISO2-M and ORB-SLAM  are geometry-based, while the others are learning-based approaches. $t_{rel}$: average translational RMSE drift ($\%$) on a length of 100-800 m. $r_{rel}$: average rotational RMSE drift ($^{\circ}/100 m$) on a length of 100-800 m.}
    \begin{tabular}{lcccc}
    \toprule
        \multirow{2}{*}{Methods}\footnotemark                              &   \multicolumn{2}{c}{Seq. \#09}    & \multicolumn{2}{c}{Seq. \#10}  \\ \cmidrule{2-3} \cmidrule{4-5}
        & $t_{rel}$       & $r_{rel}$       & $t_{rel}$          & $r_{rel}$   \\\midrule
        VISO2-M~\cite{viso2-m}         & \textbf{4.04}          & 1.43          & 25.2             & 3.8  \\
        ORB-SLAM$^{\dag}$~\cite{mur2015orb}        & 15.3                   & \textbf{0.26} & \textbf{3.71}             & \textbf{0.3} \\\hline
        DeepVO$^{\ast \dag}$~\cite{wang2017deepvo}          &  N/A                   & N/A           & 8.11             & 8.83 \\
        Wang et al.$^{\ast \dag}$~\cite{8968515}     &  8.04                  & 1.51          & 6.23             & 0.97 \\
        TartanVO$^\ddag$~\cite{tartanvo2020corl}       &  6.00                  & 3.11          & 6.89             & 2.73 \\
        UnDeepVO$^{\ast}$~\cite{li2018undeepvo}       &  N/A                   & N/A           & 10.63            & 4.65 \\
        GeoNet$^{\ast}$~\cite{yin2018geonet}         &  26.93                 & 9.54          & 20.73            & 9.04 \\ \hline
        COMPASS$^\ast$        &  \textbf{2.79}         & \textbf{0.98} & \textbf{2.41}     & {1.00} \\ \bottomrule
    \end{tabular}
    
    \label{tab:kitti-compare}
\end{table}
\footnotetext{ $t_{rel}$: average translational RMSE drift ($\%$) on a length of 100-800 m. $r_{rel}$: average rotational RMSE drift ($^{\circ}/100 m$) on a length of 100-800 m. $\ast$: the methods are trained or finetuned on the KITTI dataset. $\dag$: the methods use multiple frames to optimize the trajectory after the VO process. $\ddag$: the method is trained on large amount of labeled data.}

\begin{figure}
    \centering
    \begin{subfigure}[t]{0.48\columnwidth}
        \centering            
            \includegraphics[width=\textwidth]{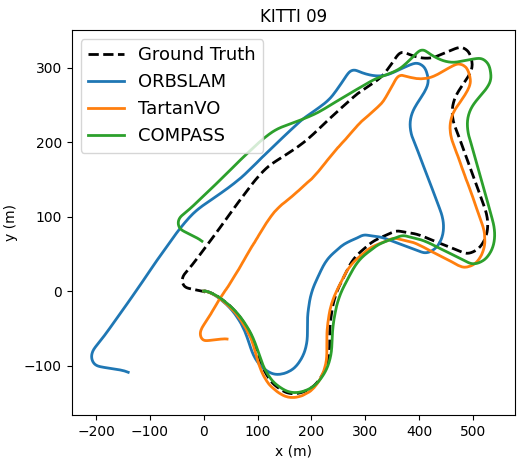}
            \caption{}
    \end{subfigure}
    \quad
    \begin{subfigure}[t]{0.42\columnwidth}
        \centering            
            \includegraphics[width=\textwidth]{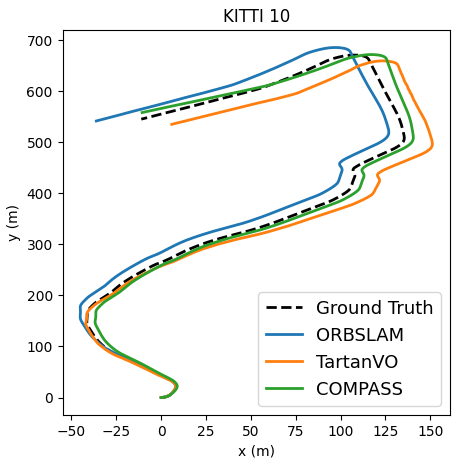}
            \caption{}
    \end{subfigure}
    \quad
    \caption{Comparison of the KITTI 9 and 10 trajectories predicted by different approaches. TartanVO~\cite{tartanvo2020corl} is a learning-based VO (only relies on two frames, same as ours), and ORBSLAM2~\cite{mur2015orb} is a geometry-based SLAM system (includes multi-frame optimization). }
    \label{fig:traj-kitti}
\end{figure}

\begin{figure*}[!t]
        \centering
        \begin{subfigure}[t]{0.23\textwidth}
            \centering            
            \includegraphics[width=\textwidth]{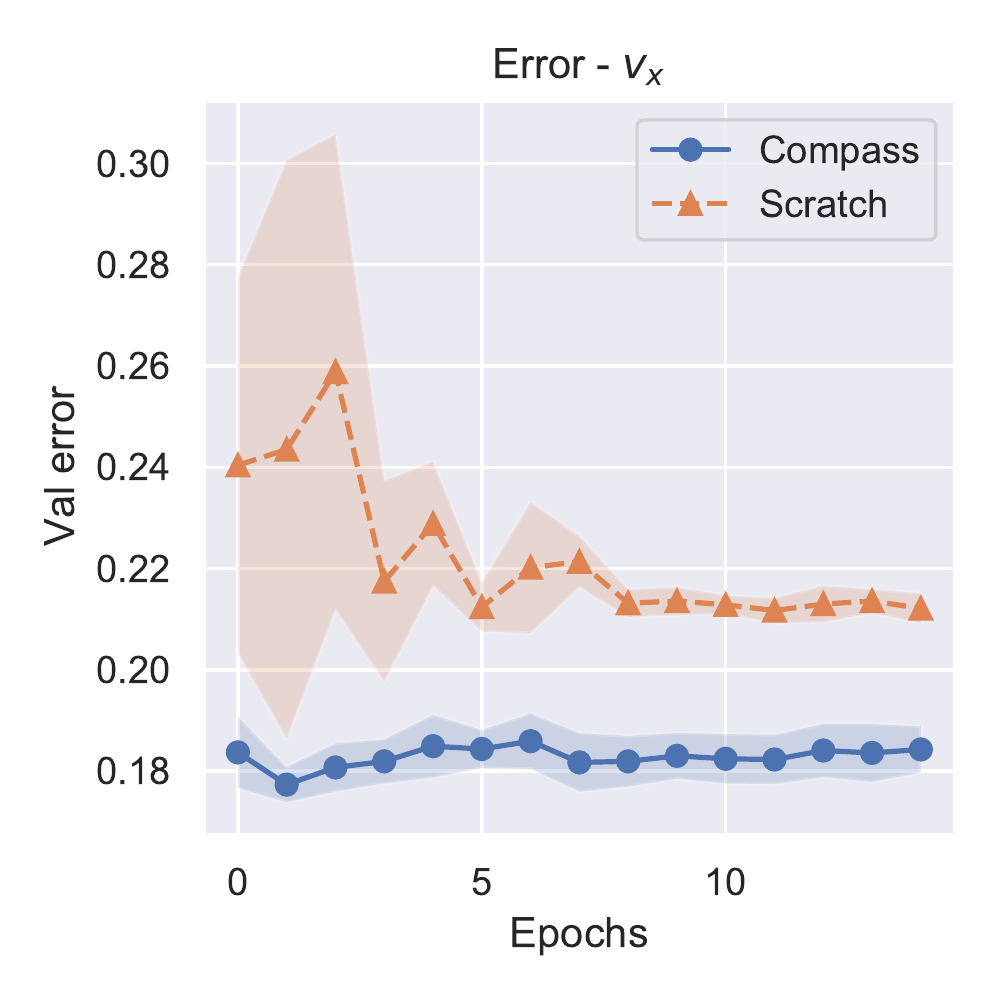}
            \caption{}
        \end{subfigure}
        \quad
        \begin{subfigure}[t]{0.23\textwidth}
            \centering            
            \includegraphics[width=\textwidth]{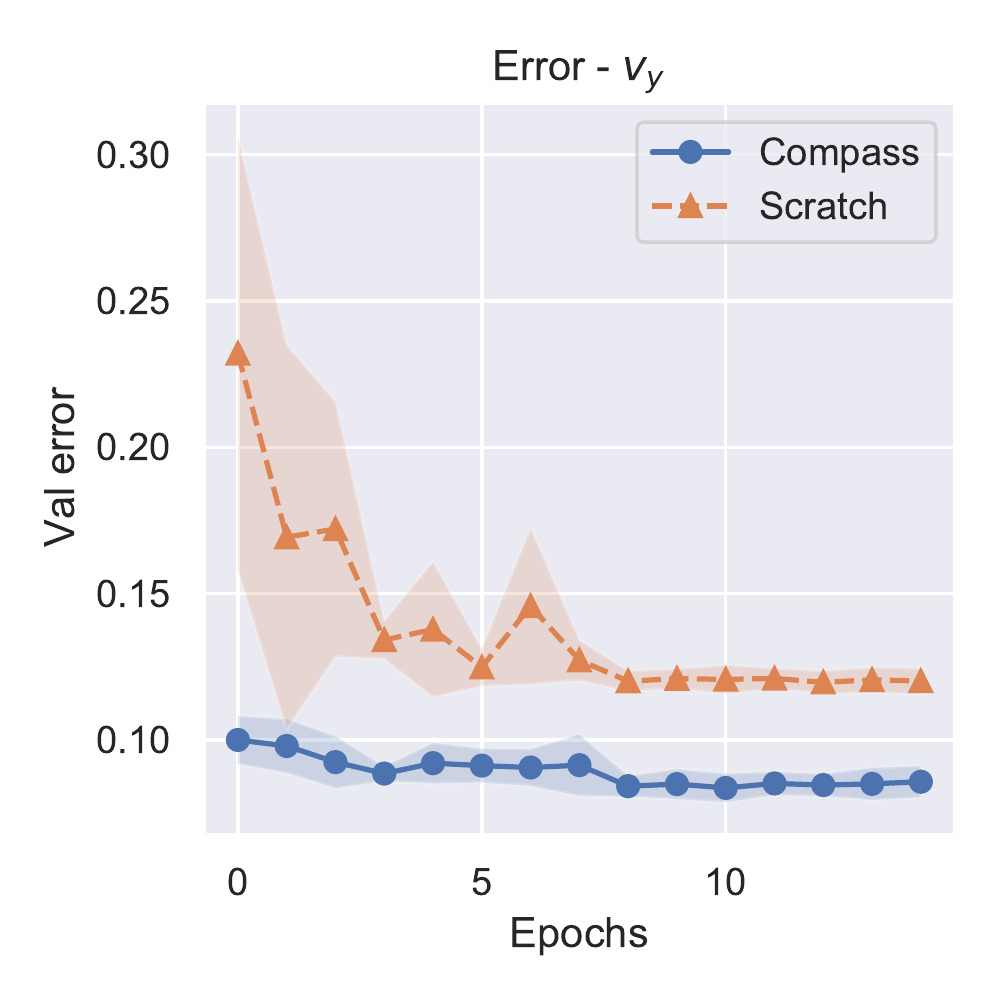}
            \caption{}
        \end{subfigure}
        \quad
        \begin{subfigure}[t]{0.23\textwidth}
            \centering            
            \includegraphics[width=\textwidth]{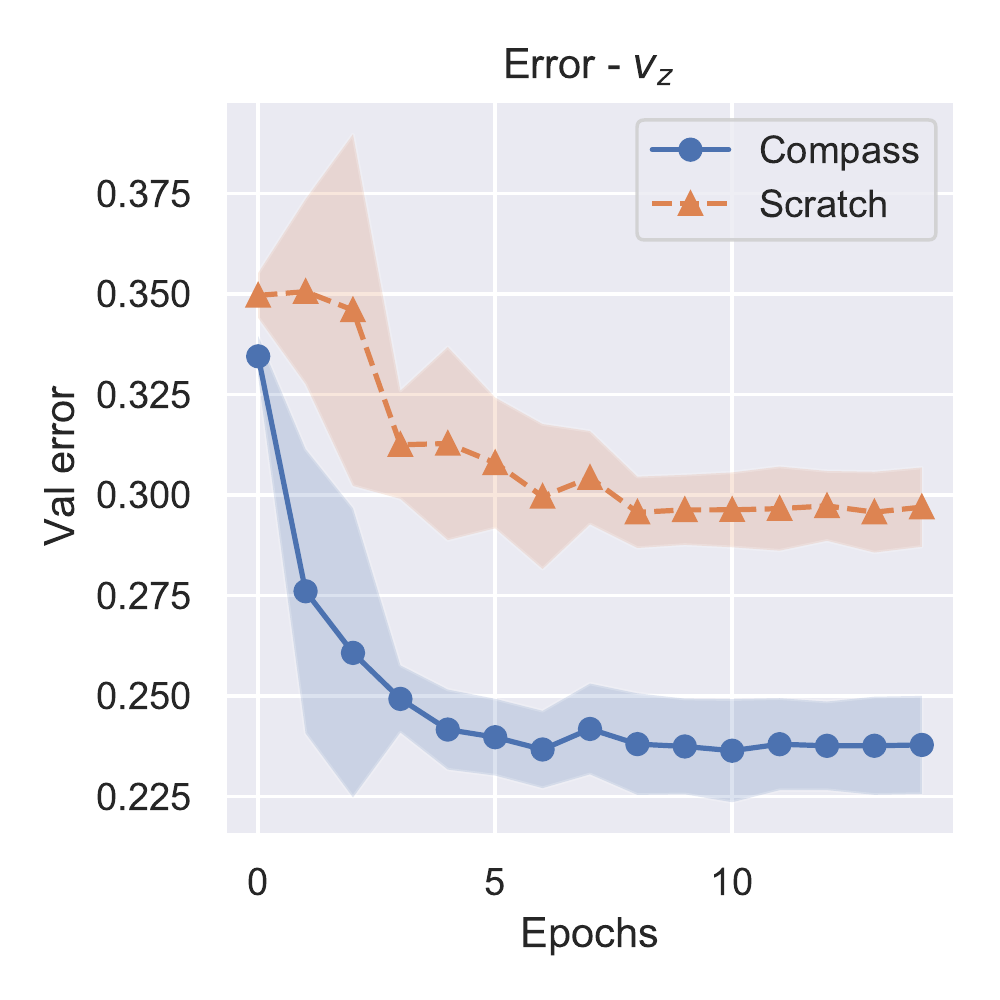}
            \caption{}
        \end{subfigure}
        \quad
        \begin{subfigure}[t]{0.23\textwidth}
            \centering            
            \includegraphics[width=\textwidth]{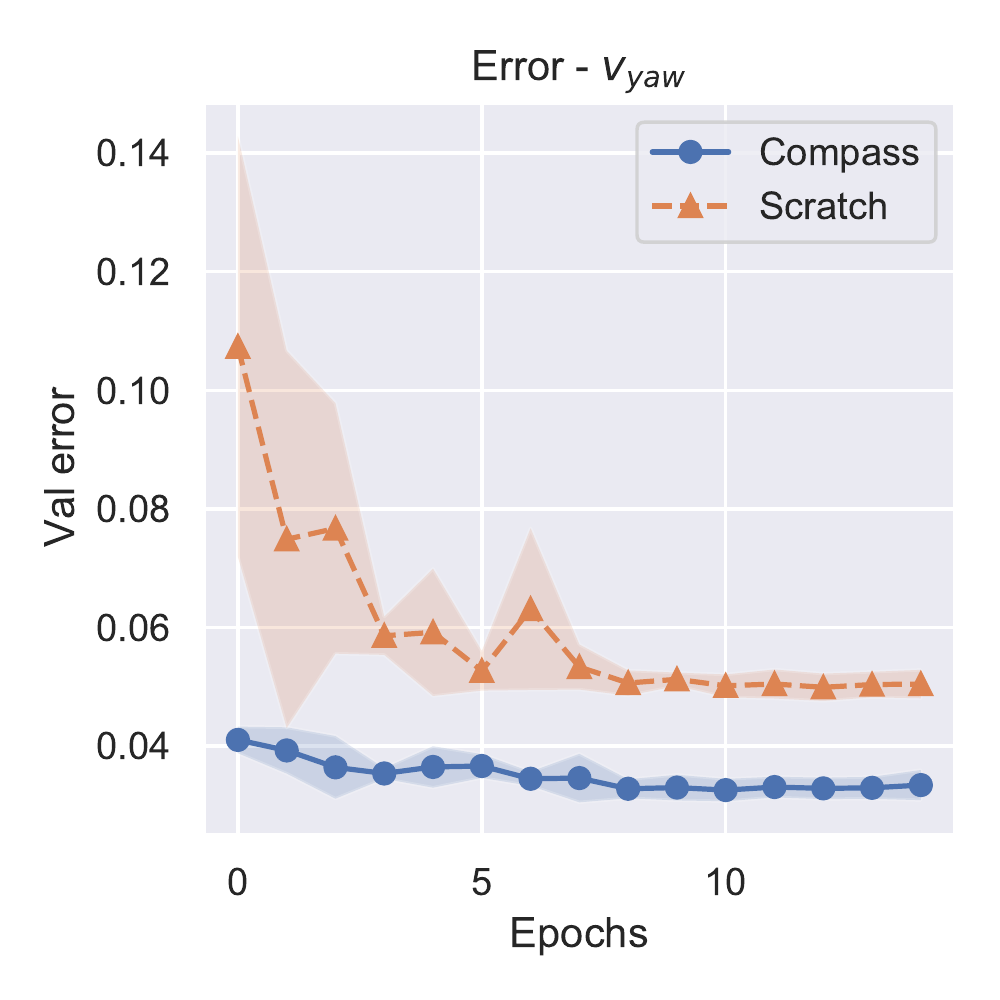}
            \caption{}
        \end{subfigure}
        \quad
    \caption{Comparison of validation error profiles of velocity predictions between COMPASS and a model trained from scratch.}
    \label{fig:drone_val_error}
\end{figure*}

\textbf{Can COMPASS be adapted to real-world scenarios?}
In this experiment, we finetune the model on sequences 00-08 of KITTI and test it on sequence 09 and 10. For comprehensive investigation, we compare COMPASS with both SLAM methods and visual odometry methods. The results are shown in Table \ref{tab:kitti-compare} using the relative pose errors (RPE) on rotation and translation, which is the same metric used on KITTI benchmark. 
The first three~\cite{mur2015orb, yang2018deep,yang2020d3vo} are SLAM methods that perform optimization on multiple frames to optimize the trajectory. 
Other baselines are VO methods that do not contain backend optimization. We compared against both geometry-based VO (VISO2-M~\cite{viso2-m}) and learning-based VO, including supervised methods (DeepVO~\cite{li2018undeepvo}, Wang et. al~\cite{wang2017deepvo}, TartanVO~\cite{tartanvo2020corl}) and unsupervised methods (UnDeepVO~\cite{li2018undeepvo}, GeoNet~\cite{yin2018geonet}). 

Using the pretrained flow encoder from COMPASS within this VO pipeline achieves results better than the other VO methods and even comparable to SLAM methods.
\Cref{fig:traj-kitti} shows the predicted trajectories of sequences 09 and 10 compared to ground truth. For clarity, we also select one representative model from the geometry-based and learning-based approaches each. We can see that, when pretrained purely on simulation data, COMPASS adapts well to real-world scenarios. 


\subsection{Drone Navigation} 
\label{sec:drone-navigation}
\textbf{Task and setting.} The goal of this task is to enable a quadrotor drone to navigate through a series of gates whose locations are unknown to it a priori. 
It takes as input monocular RGB images and predicts the velocity command. We construct the module by adding a policy network on top of the RGB encoder from pretrained COMPASS. We use a dataset from a simulated environment in AirSim Drone Racing Lab \cite{madaan2020airsim} where the environment contains a diverse set of gates varying in shape, sizes, color, and texture, through which a drone is flown to obtain the velocity commands.  

In \Cref{fig:drone_val_error}, we compare the validation errors of the $x$, $y$, $z$, and $yaw$ velocity predictions from a finetuned COMPASS model with those of a model trained from scratch. We can observe that finetuning COMPASS for this velocity prediction task results in better performance than training a model from scratch, and reaches low errors quicker than the scratch model. 

\begin{figure}
    \centering
    \includegraphics[width=0.6\columnwidth]{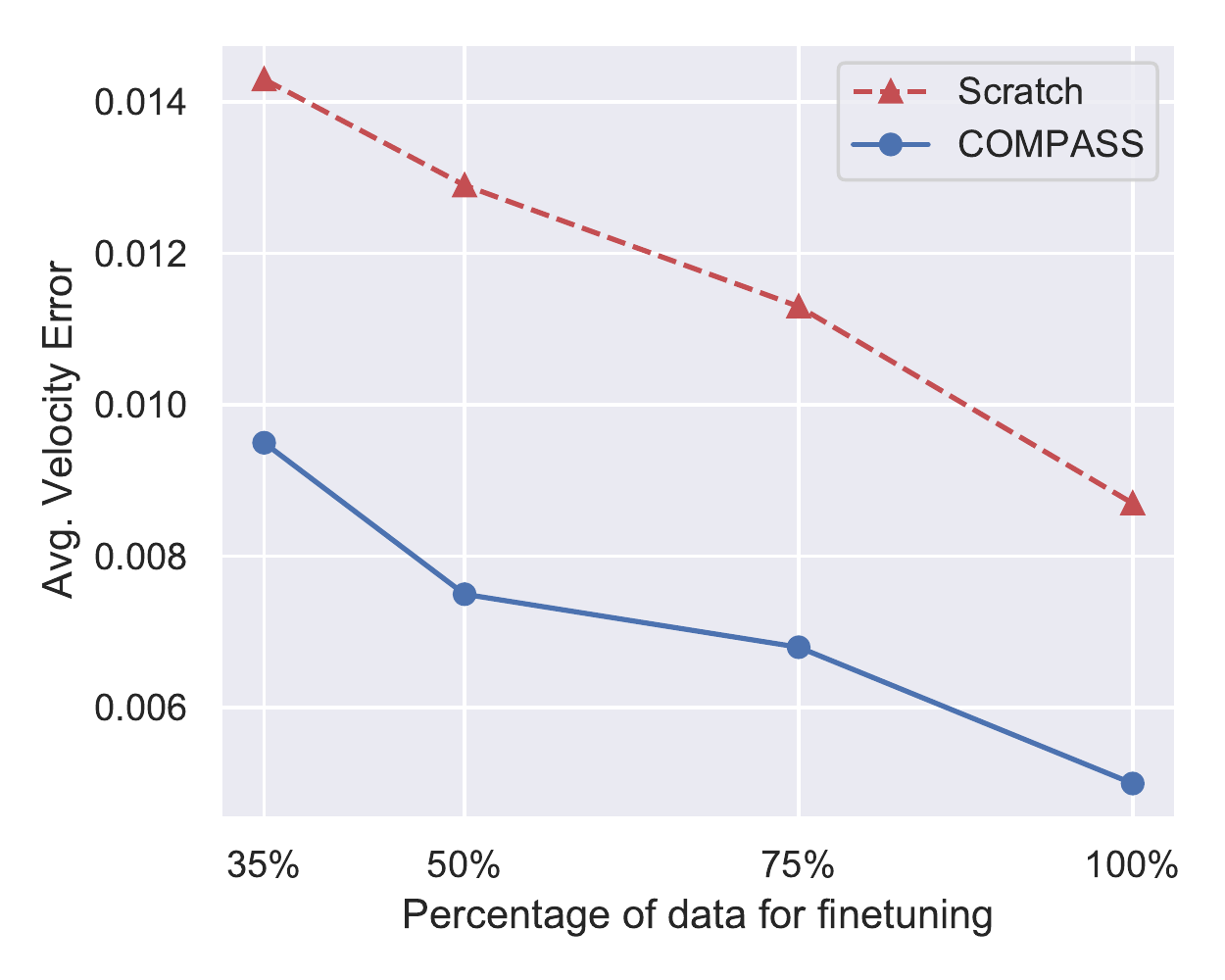} 
    \caption{Comparison of average velocity errors between COMPASS and Scratch with varying sizes of the finetuning dataset. We see that COMPASS consistently reaches a lower error even in the low-data regime.}
    \label{fig:drone-data-sample}
\end{figure}

\textbf{Can COMPASS improve data efficiency?} 
Furthermore, we observe that finetuning pretrained COMPASS models exhibits more data efficient learning than training models from scratch. \Cref{fig:drone-data-sample} compares finetuning performance with different amounts of data to training from scratch. As we reduce the amount of training samples, model finetuned on \texttt{COMPASS} generalize significantly better than the model trained from scratch. Also, to match the performance of a \texttt{Scratch} model trained on 100$\%$ data, the model finetuned on \texttt{COMPASS} only requires around 40$\%$ of the data. 

\section{Conclusion} \label{sec:conclusion}
We introduce COntrastive Multimodal Pretraining for AutonomouS Systems (COMPASS), a general purpose pretraining approach that learns multimodal representations for various downstream autonomous system tasks. In contrast to existing task-specific approaches in autonomous systems, COMPASS is trained entirely agnostic to any downstream tasks, with the primary goal of extracting information that is common to multiple tasks. COMPASS also learns modality specific properties, allowing it to encode the spatio-temporal nature of data commonly observed in autonomous systems. We demonstrate that COMPASS generalizes well to different downstream tasks -- vehicle racing, visual odometry and drone navigation -- even in unseen environments, real-world environments and in the low-data regime.

\textbf{Limitations.} More work is required for improving the local geometric information present in the learned representations. Similarly while this work was focused on using the contrastive objective to identify shared information across modalities, future work should focus on distinguishing complementary information as well. Extending the ideas presented here to even more data modalities, especially with different sampling rates and understanding scaling laws \cite{kaplan2020scaling} for such models is also important future work.


\subsection*{Acknowledgements}
We would like to thank Tom Hirshberg and Dean Zadok for their help with the vehicle racing dataset. We would also like to thank Weijian Xu for helping generate data for the drone navigation task.

\balance
\begin{small}
\bibliography{root}
\bibliographystyle{unsrt}
\end{small}

\end{document}